\documentclass[conference]{IEEEtran}
\IEEEoverridecommandlockouts

\usepackage[ruled,vlined]{algorithm2e}
\usepackage{pifont}
\usepackage{times}
\usepackage{cite}
\usepackage{amsmath,amssymb,amsfonts}
\usepackage{algorithmic}
\usepackage{graphicx}
\usepackage{textcomp}
\usepackage{booktabs}
\usepackage{xcolor}
\usepackage{multirow}
\usepackage[flushleft]{threeparttable}
\usepackage{soul,color}

\def\BibTeX{{\rm B\kern-.05em{\sc i\kern-.025em b}\kern-.08em
    T\kern-.1667em\lower.7ex\hbox{E}\kern-.125emX}}
\begin{document}

\title{DASHA: Decentralized Autofocusing System with Hierarchical Agents}

\author{\IEEEauthorblockN{Anna Anikina}
\IEEEauthorblockA{\textit{Skoltech} \\
Moscow, Russia \\
}
\and
\IEEEauthorblockN{Oleg Y. Rogov}
\IEEEauthorblockA{\textit{Skoltech} \\
Moscow, Russia \\
}
\and
\IEEEauthorblockN{Dmitry V. Dylov}
\IEEEauthorblockA{\textit{Skoltech} \\
Moscow, Russia \\
d.dylov@skoltech.ru}
}

\maketitle

\begin{abstract}
State-of-the-art object detection models are frequently trained offline using available datasets, such as ImageNet: large and overly diverse data that are unbalanced and hard to cluster semantically. This kind of training drops the object detection performance should the change in illumination, in the environmental conditions (\textit{e.g.}, rain), or in the lens positioning (out-of-focus blur) occur. We propose a decentralized hierarchical multi-agent deep reinforcement learning approach for intelligently controlling the camera and the lens focusing settings, leading to a significant improvement beyond the capacity of the popular detection models (YOLO, Faster R-CNN, and Retina are considered). The algorithm relies on the latent representation of the camera’s stream and, thus, it is the first method to allow a completely no-reference tuning of the camera, where the system trains itself to auto-focus itself.
\end{abstract}

\begin{IEEEkeywords}
Computational Cameras \& Optics, Computational Photography, Imaging \& Video; Artificial Intelligence, Machine Learning, Optimization, Methods \& Applications; Hardware\footnote{This work was supported by the Skoltech-MIT NGP Program (Skoltech-MIT joint project).}
\end{IEEEkeywords}

\maketitle
\section{Introduction}
\begin{figure}
    \centering
    \includegraphics[width=0.4\textwidth]{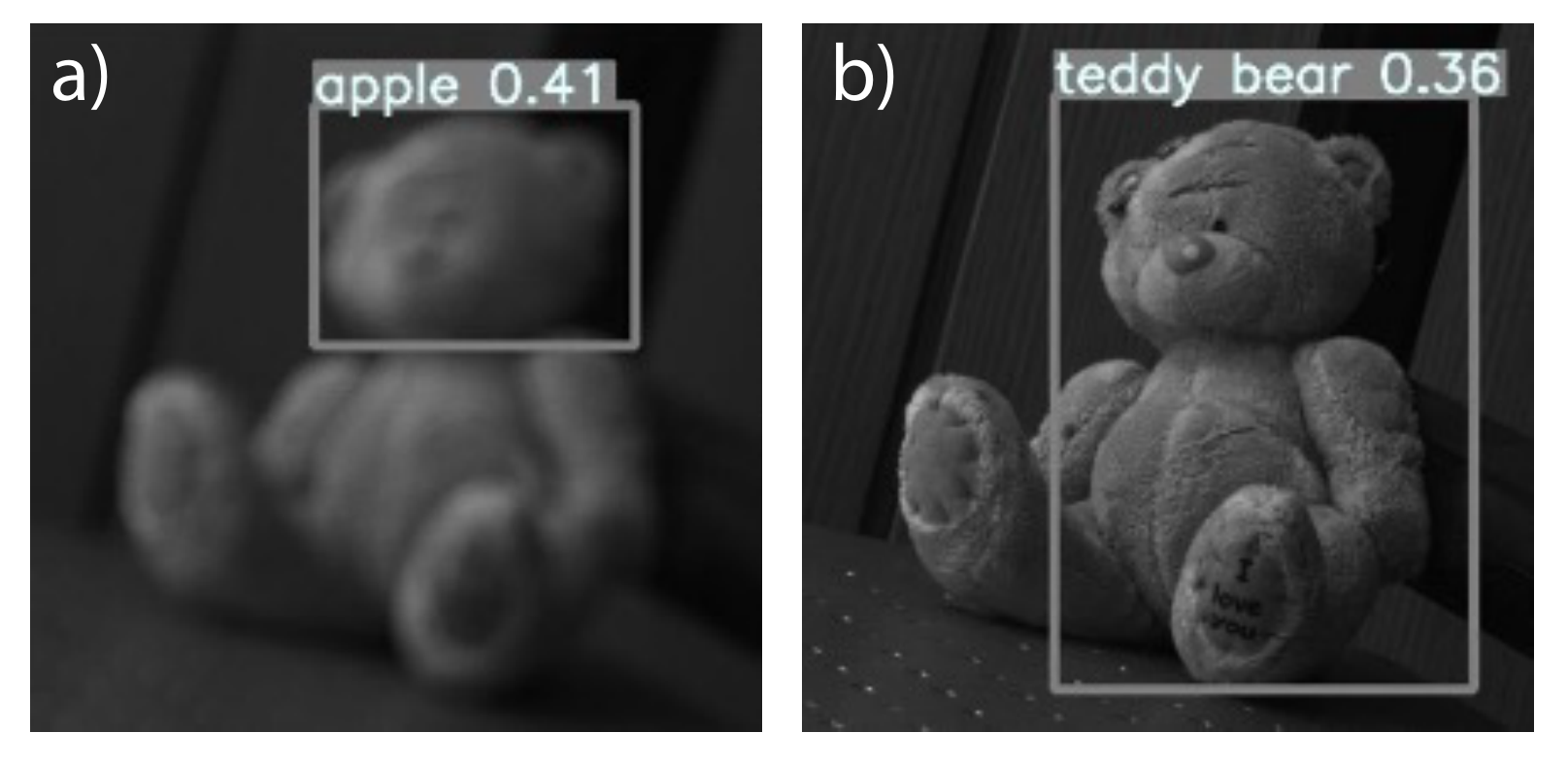}
    \caption{Motivation behind this work. Detection models fail to recognize the object in a defocused image (a). Our algorithm learns to adjust the camera and re-focuses its lens until the image looks good (b). The method requires nothing but the latent vector from the image (no reference autofocusing).}
    \label{fig:promo}
\end{figure}
Modern computational photography is unimaginable without the image acquisition hardware and image analytics tandem\cite{lukac2017computational}. 
The discipline absorbs the best of both worlds by embedding the state-of-the-art computer vision algorithms into the fastest image processing chips \cite{Scheirer_2020_TPAMI}. 
This tandem ultimately gave birth to the paradigm of $\varepsilon$-photography \cite{Raskar2009Epsilon}, where a stream of photos could be processed and computationally analyzed on the fly to overcome a wide range of camera limitations. Both limitations, optical (\textit{e.g.}, the lens’ limitations) and electronic (\textit{e.g.}, the shot noise), can be eliminated by virtue of pre-trained artificial neural networks (ANN), embedded into the camera's hardware. That enabled a plethora of photo enhancement options for the end-user, unimaginable with a single-exposure camera some 10 years ago\footnote{Such algorithmic `improvements' of the camera's hardware include the higher dynamic range, the larger depth of focus, the broader color gamut, the wider/panoramic shooting, the night photography, \textit{etc}.}.

The opposite direction of enrichment in this tandem has been inexplicably underestimated by the community until very recently \cite{Atsushi-Epsilon}. 
In particular, the embedded algorithms rarely use the arsenal of image-improving hardware components within the camera to adjust/update \textit{themselves}. 
The embedded models are typically pre-trained on offline datasets and, at best, use the recently proposed paradigm of online learning \cite{gepperth2016incremental} to update the pre-trained ANNs weights, yet, without venturing into the feedback dialogue with the camera hardware.

Recent rapid developments in optical hardware have provided an opportunity for a new approach in lens autofocusing and exposure control in the image with the aid of feedback-based control loops. Practical applications of automated camera calibrations span from on-the-fly segmentation of road scenes for autonomous vehicles~\cite{Roadseg} and smartphones~\cite{Abuolaim_2018}, to lithography~\cite{ren2018autofocusing}, to biomedical imagining~\cite{HUNTER201078,NIR}, becoming indispensable in these knowledge domains.

In this study, we were motivated by the recent advances in reinforcement learning (RL) that provide a framework for filling the niche \cite{RLAF:Chan, RLAF:Robotic}. We aspired to check if we could train \textit{hardware} in the imaging system to perform image adjustments \emph{live} in order to improve the performance of popular embedded scene analysis models, such as Yolo\cite{bochkovskiy2020yolov4}, Faster R-CNN\cite{ren2016faster}, or RetinaNet\cite{RetinaNet}.
In short, we present a perception-inspired automatic focusing system that requires no reference image and is supported by hierarchical RL. 

\medskip
The contribution of this study is in the following:
\begin{itemize}
    \item We introduce a new paradigm for passive lens autofocusing by means of hierarchical reinforcement learning;
    \item The proposed approach is decentralized: both the camera and its lens are controlled by separate agents, preserving the hardware from overheating and guaranteeing functionality in the low illuminance conditions;
    \item To the best of our knowledge, this is the first no-reference autofocusing system based on latent space representation of a live scene.
\end{itemize}

\section{Related Work}\label{s:related}

\subsection{Background}\label{s:background}

To automatically focus on an object, the exact distance from the focal plane to the object should be determined, which could be done by active and passive autofocus (AF). 

Active methods include the presence of additional (auxiliary) elements, such as, for example, an ultrasonic locator~\cite{ActiveAF:Ultrasonic}, infrared LED~\cite{ActiveAF:LED} or lasers~\cite{ActiveAF:Laser}.

Passive AF works through the analysis of the image information received by the optical camera system. Passive AF is based on two different approaches: contrast and phase detection ones. Contrast-based AF search method occurs by determining the image sharpness. To find the lens position, this approach requires capturing a sequence of images with different focal lengths and then calculating each image's Focus Measure Value~\cite{Pertuz_2013, PassiveAF:Mycobacterium, PassiveAF:Chen, PassiveAF:Xu, PassiveAF:RudiChen, PassiveAF:Zhang}. This sequence significantly affects the performance time. Phase detection methods represent the image being divided into right and left pixels on the camera sensor~\cite{PassiveAF:Gluskin, PassiveAF:Hsu}. These pixels are then compared with each other. The left and right sub-images should be similar to the lens position in focus. However, phase shifts are very sensitive to the noise, making it challenging to find the focal length~\cite{PassiveAF:Chan, RLAF:Chan, PassiveAF:Gluskin2}. 

Recently, many articles regarding the use deep learning techniques for the AF task ~\cite{DeepAF:Herrmann}, including super-resolution techniques were published ~\cite{DeepAF:Wang}. As a rule, an image is fed to the input, and the network predicts the focal length. Some methods approach the AF problem from the depth maps perspective ~\cite{DeepAF:Eigen, DeepAF:Carvalho, DeepAF:Garg, DeepAF:Wang}. Another promising approach features per-pixel depths obtained from multi-view stereo~\cite{L2AF}.

\subsection{Autofocus with Reinforcement Learning}\label{s:af_rl}
Surprisingly few articles cover the reinforcement learning (RL) approach to the AF problem. These works use the reference value to evaluate the resulting image, subsequently imposing serious restrictions on a hardware-based system. Therefore, given a variable ambient environment the computations are dramatically time-consuming and are unfavorable for the cutting-edge applications in photography.

The work in~\cite{RLAF:Chan} appears to be the only approach that applies RL to the AF directly. There are three essential parameters in RL: the action space, the observation space, and the reward function (see:~\ref{sec:RL}) that an agent interacts with. The action space in~\cite{RLAF:Chan} is the lens movement; the observation space is the phase shift and the lens travel distance at the last time step. The reward function consists of the ground-truth value of the lens distance to the in-focus value, thus imposing a restriction on the agent training process. As the distance changes, one has to stop the training procedure each time to change the parameters  environment and retrain the agent once again. Since the approach uses empirical coefficients for the specific sensor, the  realized approach cannot be extended to other cameras. 

The article in~\cite{RLAF:Robotic} features a deep reinforcement learning AF method for the microscope lens control. The action space is a multi-discrete one and an image is taken as the observation space. The reward function is calculated using the Tenengrad algorithm (TEN)~\cite{Tenengrad}. It also uses a reference value: during the initialization of the RL environment, the search function for the focal threshold value is called. It iterates over all the lens values and calculates the maximum focal value from the resulting images. Such a system is not resistant to the ambient environment changes such as illumination or object movements since the latter would have a different reference value. In this case inside the RL environment one has to call the threshold value check function once at N time steps. The strength of the work points out the fact that a robot hand is used to adjust the focus, which controls the microscope's settings. However, if one transfers this method to a purely camera setup, there would be no need for RL as the statistical methods are employed to find the focal length.

\subsection{Exposure adjustment with Reinforcement Learning}\label{s:exp_rl}

Besides, a small number of articles were found covering the exposure time. All these articles do not use direct learning on the hardware.

In \cite{ExTime:Yang} the authors approached the prediction of exposure time $t_{ex}$ as a function of the current frame. In order to train the network to automatically adjust exposure times, the authors use a “coarse-to-fine” learning strategy. This strategy consists in the the network pre-training of the network on the collected dataset to predict the $t_{ex}$ (coarse training). Only after this action an additional training occurs online using RL (fine training). This approach is used due to the large time spent on training the RL. An image comes from hardware (with configured $t_{ex}$ from the backbone network), and users rate this image following 3 main categories: “under-exposed,” “correctly exposed,” or “over-exposed”. Based on these 3 categories, the reward is either -1 or 1.

The article in \cite{ExTime:Yu} describes an approach to setting the exposure time: an image is taken (from a ready-made dataset), the areas are segmented, and then an action is applied to each area that either darkens or lightens the segment. At the end, all areas are merged again, and the final image is thus obtained. As a reward, a piecewise function is used: 0 for each action within the episode, and the aesthetic evaluation function of the final image at the end of the episode. In order to obtain the function of aesthetic assessment, the authors use a generative adversarial network (GAN). The function is a discriminator and aims to evaluate the image produced by the RL from the image that was processed by the expert. This function only works with the dataset previously used for training.

\section{Methods}

\begin{figure*}
    \centering
    \includegraphics[width=\textwidth]{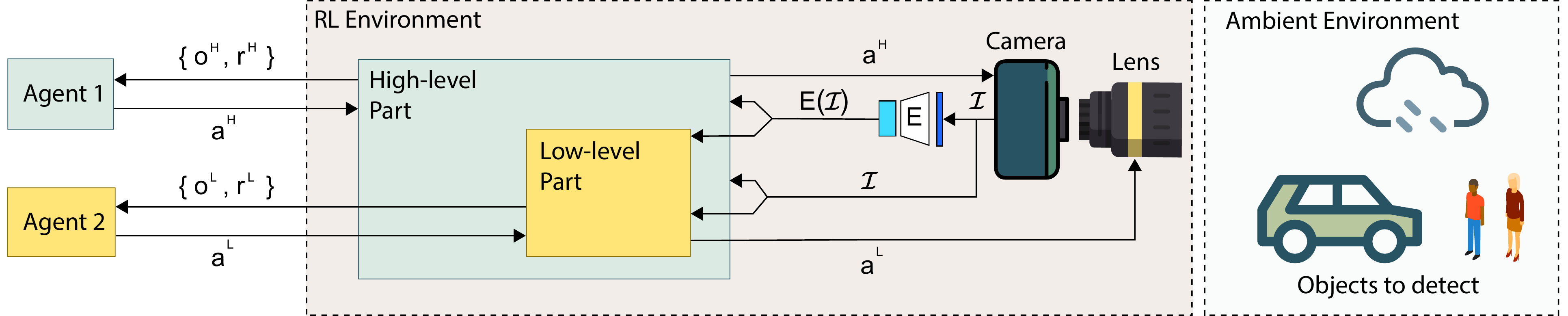}
    \caption{Decentralized autofocusing system with hierarchical RL agents (DASHA). There are two agents operating in the RL-Environment: a high-level agent (Agent 1: turquoise) and a low-level agent (Agent 2: yellow). $o, r, a$ - observation, reward, and agent's action, respectively. The superscripts $H$ and $L$ indicate the high-level and the low-level agents, respectively; $\mathcal{I}$ - image obtained from camera; $E$ is the pre-trained ResNet-152 encoder and E($\mathcal{I}$) - the feature vector.}
    \label{fig:masterfigure}
\end{figure*}

\subsection{Reinforcement learning}
\label{sec:RL}
Consider RL problem as a Markov Decision Process (MDP) formally represented in form of tuple $\left(\mathcal{S}, \mathcal{A}, \mathcal{P}, \mathcal{R}, \gamma, \right)$, where $\mathcal{S}$ is a finite set of states and $s_t \in \mathcal{S}$ denotes that agent at time $t$ being in state $s$; $\mathcal{A}$ is a set of actions and agent interacts with the environment by choosing action $a_t \in \mathcal{A}$ at time t; $\mathcal{P} : \mathcal{S} \times \mathcal{A} \times \mathcal{S} \rightarrow \mathbb{R}$ is the transition probability distribution describes the probability of arriving from $s_{t-1}$ to $s_t$ by choosing $a_{t-1}$; $\mathcal{R}$ is a scalar reward function that controls the behavior of agent: the environment gives reward to agent for each action; and $\gamma \in [0, 1]$ is the discount factor providing the importance of immediate reward versus future rewards.

The agent $A$ starts from a state $s_t$ at time $t$, choose an action $a_t$, obtains a reward $r_t \in \mathcal{R}$ and new state $s_{t+1}$. This sequences of action repeat until agent reaches a success state or/and end of the episode. The number of steps in each episode is the horizon ($H$). Thus, an episode of a MDP is represented as a sequence $(s_t,a_t,r_t,s_{t+1},a_{t+1},r_{t+1},s_{t+2},a_{t+2},r_{t+2},...)$. For MDP there is a policy $\pi: \mathcal{S} \rightarrow \mathcal{A}$ -- a function that maps the agent's states to actions and defines action to take in each state. It can be  represented by a tuple $\pi = (\pi_t, \pi_{t+1}, \pi_{t+2}, ...)$, where $\pi_t$ refers to the policy at time $t$. The objective is to find an optimal policy $\pi$ by maximizing the future rewards.

At this stage, two important concepts should be introduced: value and state value functions. The state value function $V^{\pi}$ is the cumulative sum of future rewards obtained by the agent following the policy $\pi$ from the state $s$: 
\begin{equation}\label{eq:valuefun}
V^{\pi}_t(s) = \mathbb{E}\left[\sum^{H-1}_{i = t} \gamma^{i-t}r_i|s_t = s\right]
\end{equation}

The state-action value function $Q^{\pi}_t$ defines the expected discounted sum of rewards $\sum_{t=0}^{N}r_{t}$ from state $s$ at time $t$ to the $H$ following the policy $\pi$:

\begin{equation}\label{eq:state_action_fun}
Q^{\pi}_t (s, a) = \mathbb{E}\left[\sum^{H-1}_{i = t} \gamma^{i-t}r_i|s_t = s, a_t = a\right]
\end{equation}

The $V^{\pi}$ provides the best value following policy $\pi$ for each of next states, while $Q^{\pi}$ shows the effectiveness of actions that the agent chooses by following the policy $\pi$ for next states. 

There are two approaches in the model-free RL methodology (without a transition probability distribution): the value one and the policy-based one. In the first case, the state-action value function is determined and then -- the extract policy. In the second case, the policy is optimized directly.

\paragraph*{Proximal Policy Optimization} As the PPO is a policy-based method, consider a $L(\theta)$ as an objective function required to be maximized over the policy parameters $\theta$. 

The policy gradient theorem\cite{Sutton_NIPS99} states: for any differentiable policy $\pi_{\theta}(a|s)$ and for any policy objective functions, the policy gradient is

\begin{equation}\label{eq:objfuntheorem}
\nabla_{\theta} L(\theta) = \mathbb{E}_{\pi_\theta}\left[ Q^{\pi_\theta} (s, a) \nabla_{\theta} \log \pi_\theta (a|s)  \right]
\end{equation}

And from the variation of policy gradient theorem~\cite{RL:optim} the objective function is defined as follows:

\begin{equation}\label{eq:gradient_th}
 L(\theta) = \hat{\mathbb{E}}_{t} \left[\log \pi_{\theta} (a_t|s_t) \hat{A_t} \right]
\end{equation}
where $\hat{\mathbb{E}}_{t}$ denotes the empirical expectation over time steps, $\hat{A}_{t}$ is the estimated advantage at time $t$:

\begin{equation}\label{eq:advantage}
\hat{A}_{t} := Q^{\pi}_t (s, a) - V^{\pi}_t (s)
\end{equation}

Using the Trust Region Policy Optimization (TRPO)~\cite{RL:TPRO} that maximizes objective function subject to a constraint on the size of the policy update, ~(\ref{eq:gradient_th}) takes the following form:

\begin{equation}\label{eq:trpo}
 L(\theta) = \hat{\mathbb{E}}_{t} \left[\hat{r}_{t} \hat{A}_{t} \right]
\end{equation}
where $\hat{r}_t$ is the probability ratio under the new and old policies:
\begin{equation}
    \hat{r}_t = \frac{\pi(a_t | s_t)}{\pi_{old}(a_t | s_t)}
\end{equation}

Yet, TRPO may lead to instability due to large policy update. We therefore transform it in order to penalize large policy changes to stay within a small interval $[1 - \epsilon, 1 + \epsilon]$, where $\epsilon$ is a hyperparameter. The PPO's \cite{RL:PPO} objective function defines as:

\begin{equation}\label{eq:ppo}
L (\theta) = \hat{\mathbb{E}}_{t}\left[\min \left(\hat{r}_{t}(\theta) \hat{A}_{t}, clip\left(\hat{r}_{t}(\theta), 1-\varepsilon, 1+\varepsilon\right) \hat{A}_{t}\right)\right]    
\end{equation}
where $clip$ stands for the clipped objective version. The part with the $clip\left(\hat{r}_{t}(\theta), 1-\varepsilon, 1+\varepsilon\right)$ does not allow $\hat{r}_{t}(\theta)$ to go beyond interval $[1 - \epsilon, 1 + \epsilon]$ by modifying the surrogate objective by clipping. 

\subsubsection{Hierarchical Reinforcement Learning}\label{subs:HRL}
Hierarchical Reinforcement Learning (HRL) is the case of multi-agent RL and has multiple hierarchy layers of policies compared to the conventional single RL agent. In the standard two-structured HRL, high-level agent $A^{H}$ and low-level agent $A^{L}$ are defined with corresponding policies $\pi^{H}$ and $\pi^{L}$. The higher-level policy sets goals for the lower-level policy, and the lower-level policy attempts to reach it. At the beginning of each episode, the higher-level policy receives the state (or observation) and forms a high-level action (or goal). Then, the low-level policy receives an observation and a goal and forms a low-level action that affects the environment. The low-level agent receives a reward for each step; at the end of the episode, the high-level agent receives the final reward, and then the process is repeated. 

Our approach differs from the classic HRL. The high-level agent establishes a 'favorable' RL-environment for the low-level agent to act: until the optimal parameters of the internal environment are reached, the low-level agent does not take control. The high-level agent in our work is responsible for the camera parameters: it controls the $t_{ex}$ and adjusts the camera to a visible image. A low-level agent works with the lens: the task is to get the selected object in focus. The object is detected by the state-of-the-art (SOTA) methods described in Section~\ref{sub:objdetect}.

\subsection{Object detection}
\label{sub:objdetect}

Object detection is an essential part of the environment. The camera image itself may be partially blurred with the selected object in focus. When analyzing such an image, the metrics can show that the image is unacceptable, although there is a selected object in focus.

\subsubsection{Yolo}
The key feature of YOLO consists in CNN being applied once to the entire image, predicting multiple objects and yielding the class probabilities for the objects. YOLO divides the incoming image into an $N \times N$ grid. If the center of an object falls into a cell of such a grid, this cell is responsible for detecting the object. Each cell predicts a certain number of bounding boxes and gives a probabilistic estimate for each box: how confident the model is while this box contains an object, and how confident the model is while the object in this box belongs to a particular class. The estimation of the returned result reliability is performed by the intersection on union (IoU) between the predicted block and the ground truth~\cite{bochkovskiy2020yolov4}. 

\subsubsection{RetinaNet} 

RetinaNet is a single network of 3 essential components: a backbone network and two subnets for specific tasks. Backbone, the the main or basic network, is responsible for calculating a convolutional feature map from the input image (in our case, such a network is based on the ResNet-152\cite{He_2016_CVPR} classification neural network). Backbone also includes the Feature Pyramid Network (FPN)\cite{Lin_2017_CVPR}. Its purpose resides in combining the advantages of feature maps of the network lower and upper levels, since the lower levels have high resolution, while low have semantic ones. The two subnets are the following. The Classification Subnet is the subnet that extracts information about object classes from FPN and solves the classification problem. The Box Regression Subnet is the subnet that extracts information about the object coordinates in the image from FPN and solves the regression problem~\cite{Retina}.

\subsubsection{Faster R-CNN} 

The entire image and a set of detected objects are the input of the Faster R-CNN. First, a feature map is created - this is done with several convolutional and max-pooling layers. Then a fixed-length feature vector is extracted from the feature map for each object, this vector is passed to a sequence of fully connected layers eventually branching into two output layers: one layer gives the probability for the feature classes, and the other layer gives the object box parameters~\cite{frcnn}.

\begin{algorithm}[b]
 \caption{~~High-Level Agent}
 \begin{algorithmic}[1]
 \renewcommand{\algorithmicrequire}{\textbf{Input:}}
 \renewcommand{\algorithmicensure}{\textbf{Output:}}

  \STATE $A^H$ sends action $a^H$ 
  \STATE Change $t_{ex}$ on the camera
  \STATE Grab the image $\mathcal{I}$
  \STATE Evaluate $a^H$ of the $A^H$ by computing \texttt{histogram}
  \STATE Form observation $o^H$ from $\mathcal{I}$ using encoder $E(\cdot)$ and \texttt{histogram}
  \STATE Compute reward $r^H$
  \IF {(\texttt{histogram} is unsuitable)}
    \STATE Episode completed with a penalty for $A^H$
 \ELSE
    \STATE Pass the action to $A^L$ (Low-Level Agent) or successfully end the episode (in case of Single Agent)
\ENDIF

\RETURN ($o^H, r^H, done$)
 \end{algorithmic} 
 \end{algorithm}
 
\subsection{Algorithm}

This section covers the method proposed in Fig.~\ref{fig:masterfigure}.

\subsubsection{State Space}
A number of modern RL algorithms are still reported to solving tasks with $\mathcal{S}$ spaces of rather low dimensionality~\cite{bengio2005curse,RL_dim}.
Unlike classical computer vision approaches, we take advantage of deep auto-encoders unsupervised training, thus acquiring a low-dimensional feature space.
The $A^{L}$ observation space $s_{t} \in \mathcal{S}$ is a feature vector of size [$2048 \times 1$] from the camera image $\mathcal{I}$ obtained using a ResNet-152 Encoder $E(\cdot)$ pre-trained~\cite{Resnet} with ImageNet~\cite{imagenet}:

\begin{equation}
o^L = E(\mathcal{I})  
\end{equation}

The observation space of $A^{H}$ is a tuple of features and information from the histogram (values of the histogram $val$ [$10 \times 1$] and bin edges $bin$ [$11 \times 1$]):
\begin{equation}
o^H = (E(\mathcal{I}), val, bin)  
\end{equation}

\subsubsection{Action Space}\label{s:actionspace}
Each policy has its own set of actions. The action space for an $A^{H}$ is a discrete space of 146 values, where each value corresponds to its own exposure time. The values of the $t_{ex}$ on the camera are distributed non-linearly and selected empirically. Otherwise, the action space would consist of 5 million values, where the difference between some values would not be physically noticeable but would significantly increase the training time. The action space for $A^{L}$ is a multi-discrete space of 2 values: coarse and fine-tuning of the Focus Measure Value. The minimum value that can be applied to the lens is 24.0, and the maximum value is 70.0.

\subsubsection{Reward Function}\label{s:rewardfunc}
The $A^{H}$ gives the $A^{L}$ an intrinsic reward for each step. The reward is defined by a piecewise function that takes into account different cases of agent behavior.

To properly construct the reward function, the following points were accounted for:

\begin{itemize}
    \item  Speed. The faster the agents reach the desired state of the environment, the better. The more unsuccessful steps an agent takes, the greater is the penalty.
    \item Encouragement for approaching the border of focus. When the method detects an object, but with the result of the boxing, the speed of the image quality assessment is low and insufficient to pass the required threshold. In this case, the agent receives a small penalty, as it is generally moving in the right direction. The closer to focus, the smaller the penalty.
    \item The reward function convergence. A limit is set on the maximum that agents can receive for their actions. This prevents the model's trend to get reward irregardless of the policy.
\end{itemize}

Rewards are determined as follows. For the $A^{H}$:

\begin{equation}
  r^{H} =
  \begin{cases}
    1 &, \text{$P \in [50, 150]$ (1)} \\
    -0.01 \cdot\mathcal{B}  &, \text{$P \in [25, 50) \cup (150, 175]$ (2)} \\
    -1 &, \text{otherwise}
  \end{cases}
\end{equation}

(1) Each step of the $A^H$ is estimated by the resulting histogram from the image. If the peak $P$ of the constructed histogram falls within the interval from 50 to 150 pixel values, then the image is considered good. Otherwise, the agent receives a penalty.

(2) The image is still visible if the $P$ belongs to the interval from 25 to 50 or from 150 to 175 pixel values. In this case the action is considered partially successful and a small negative reward is received. The Blind/Referenceless Image Spatial Quality Evaluator (BRISQUE) \cite{BRISQUE} (thereafter referenced as $\mathcal{B}$) is calculated: the lower the value, the better is the image in terms of the metric ranging from 0 to 100. \textit{Literally, by ``no reference'' we mean there is no ground truth image for the agents to learn from.}

For the $A^{L}$:

\begin{equation}
  r^{L} =
  \begin{cases}
    -1  &, \text{det = False (3)} \\
    -0.01\cdot\mathcal{B} &, \text{otherwise (4)} 
  \end{cases}
\end{equation}

(3) For each step when the object was not detected, the agent receives a penalty.

(4) In the case when the object has been detected, $\mathcal{B}$ is calculated as well. Thus, the agent aims to reach the lowest possible value $\mathcal{B}$ while tending to converge to zero.

\subsubsection{High-level Agent}

This agent manages the camera settings. The main role of this agent is to set up an 'optimal environment' for the $A^{L}$. At the beginning of each episode, the camera and the lens have 'factory' initial parameters of the features. These parameters are not always optimally matched with the environment. Obtaining an optimal environment is the correct configuration of parameters so that a visible image can be obtained. The $A^{H}$ sends an action $a^{\text{H}}$ to the camera, which changes the exposure time. The resulting image from the camera $\mathcal{I}$ is analyzed for visibility by constructing a histogram. If the image is good, then further control of the environment is given to the $A^{L}$. If the image is unacceptable, the episode ends immediately with a penalty for the $A^{H}$. 

\begin{algorithm}[b]
 \caption{~~Low-Level Agent}
 \begin{algorithmic}[1]
 \renewcommand{\algorithmicrequire}{\textbf{Input:}}
 \renewcommand{\algorithmicensure}{\textbf{Output:}}
 \STATE $A^L$ sends action $a^L$ 
 \STATE Change the lens position
 \STATE Acquire the image $\mathcal{I}$
    \IF {(Object is detected)}
     \STATE Successful completion of the episode
    \ELSE
        \STATE Send new action $a^L$ until the episode is \texttt{done} 
    \ENDIF
 \STATE Form observation $o^L$ from $\mathcal{I}$ using encoder $E(\cdot)$
 \STATE Compute reward $r^L$
 \RETURN ($o^L, r^L, done$)
 \end{algorithmic} 
 \end{algorithm}
 
\subsubsection{Low-level Agent}

The $A^{L}$ is used to adjust the lens for an in-focus image acquisition. As soon as the action of the agent $a^{L}$ is transferred to the lens, a new image is acquired from the camera. This image $\mathcal{I}$ is input of the object detection model that returns the binary result: the object was successfully detected (in this case, the coordinates of the object are calculated) and the object was not detected. All these actions are repeated until the moment of successful detection or until the end  of the episode is reached. In case of successful object detection, the resulting image box is evaluated for quality using the $\mathcal{B}$. 

\begin{table}[t]
\begin{center}
\label{tab:key_measurements}
  \caption{RL-based autofocusing methods comparison: key measurements.}
   \begin{threeparttable}
  \begin{tabular}{*{15}{c}}
    \toprule
     & Phase shift\cite{RLAF:Chan} &  TEN method\cite{RLAF:Robotic} &  \textbf{DASHA} \\
    \midrule
    IQA & RB & RB & \textbf{NR}\\
    AF time (sec) & 58 & 366 & 10\\
    AF steps & 12.5 & - & 2\\
    Virtual pre-train & \ding{55} & \ding{51} & \ding{55}\\
    DF & \ding{55} & \ding{55} & \ding{51}\\ 
    \bottomrule
  \end{tabular}

      \begin{tablenotes}
      \small
      \item \textbf{Note:} IQA - Image Quality Assessment, RB - reference-based, NR - no reference, AF time - average time to set the image in focus, AF steps - the average number of steps to find the image in focus, Virtual pre-train -  a pre-trained agent before online work with hardware, DF - deep image features (latent space). \textit{For comparison, classical Phase Detection AF (PDAF \cite{Abuolaim_2018}) takes 8--14 sec under similarly low illuminance (and an extra phase element).}
    \end{tablenotes}
    \end{threeparttable}
 \end{center}
\end{table}

\section{Experiments}
For the experiments, we use the 2/3'' Basler acA2000-50gm camera (GigE, CMV2000 CMOS-matrix, 50 frames per second (FPS) at a 2 megapixels resolution). The access to the camera is established via official Basler library \texttt{pypylon}. We use the Corning Varioptic C-C-39N0-250 Lens controlled through the cp210x board (Silicon Labs Ltd.), and a DLL-file provided by Corning. To implement RL and operate both camera and lens we use the \texttt{rllib}~\cite{liang2018rllib}. We used OS Windows 10 64 bit, 16 GB RAM, CPU Core i7 3.6 GHz.

\begin{figure}[b]
    \centering
    \includegraphics[width=0.43\textwidth]{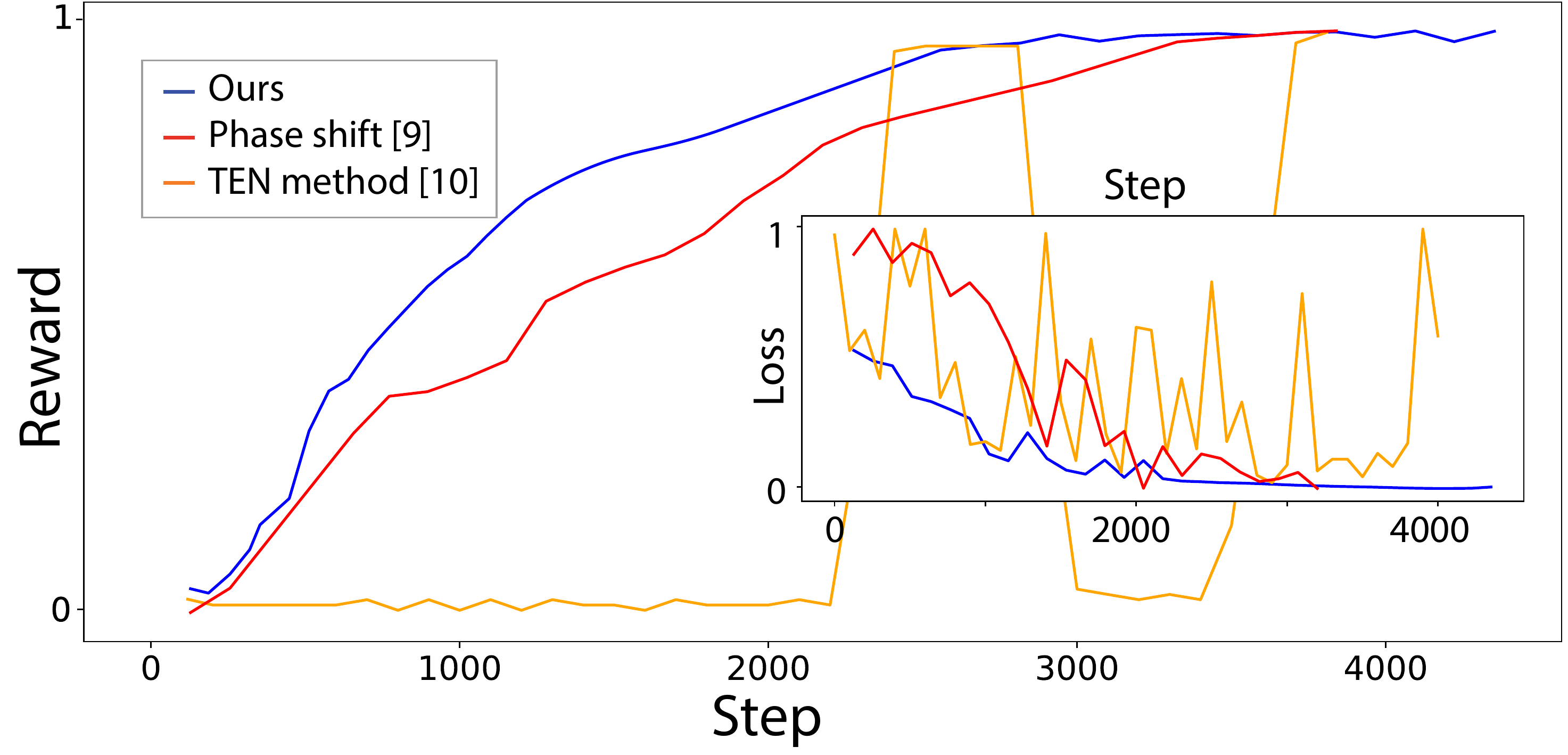}
    \caption{Learning curves of the competing autofocusing methods, demonstrating faster convergence of DASHA. Inset compares the three loss functions.}
    \label{fig:rewards_compared}
\end{figure}

\subsection{Single Agent Training}
\paragraph*{Autofocus} Prior to the multiple-agent RL training, we trained the single agent for the AF task. The agent is thus trained following the aforementioned Algorithm 2 of $A^{L}$. The learning process consists in changing the object position once every 30 minutes. 

\begin{figure*}[t]
    \centering
    \includegraphics[width=0.9\textwidth]{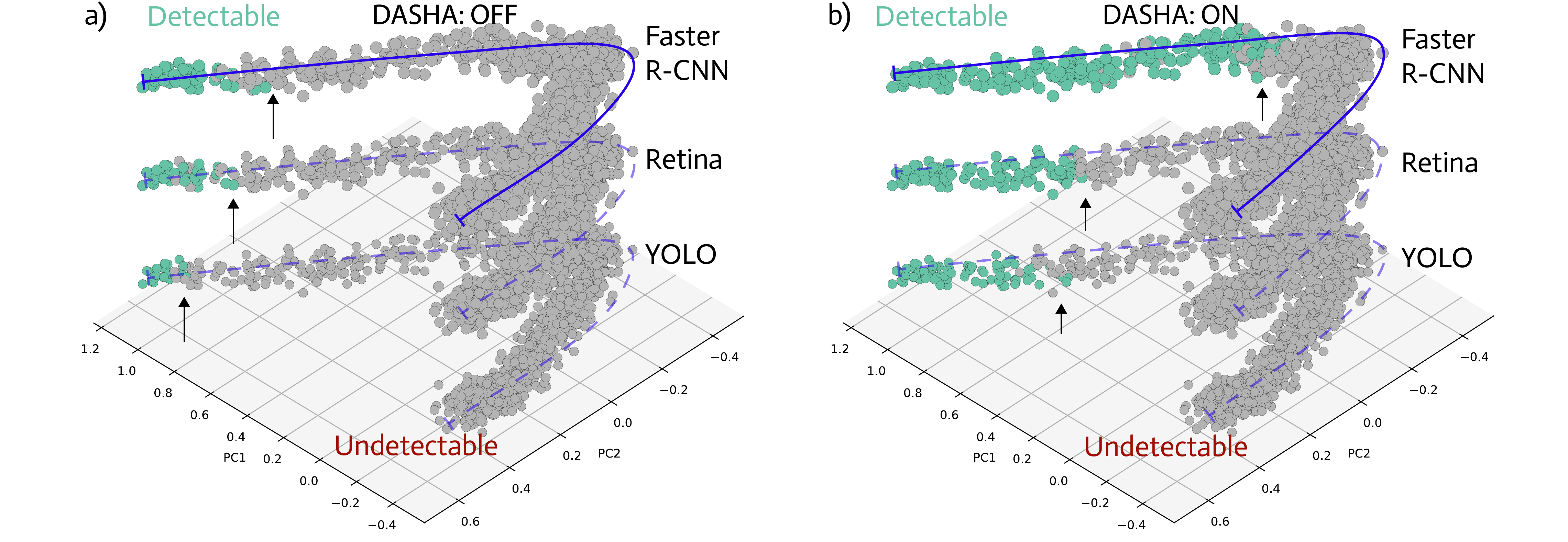}
    \caption{PCA plots demonstrating enhancement of object detection capacity of the popular models by DASHA. When DASHA is switched on, the detection boundary moves rightwards (highlighted by arrows). Each dot corresponds to an image taken at fixed illumination 37 lx and adjustable exposure time and focus. Blue lines are drawn to guide the eye.}
    \label{fig:PCA}
\end{figure*}

In order to demonstrate the advantage of the proposed agent system in the AF task, we report a comparison with~\cite{RLAF:Chan} and~\cite{RLAF:Robotic}, since these methods are the most relevant to the present study. By replicating the described reward functions and the observation space, we train the PPO agent with the results given in Table 1.

The image quality assessment (\emph{IQA}) is the method implemented to assess the acquired image within the RL-environment. The reference-based (RB) method uses a ground-truth value to compare the resulting image and the no-reference (NR) one -- without the ground-truth, respectively. \emph{AF time} ($t_{AF}$) is measured in seconds and describes the average time required for an object to get into the focus plane. 

In~\cite{RLAF:Robotic}, the time required to find the proper Focus Measure Value is obtained in the following way. An image is acquired for every lens position with a respective calculated Focus Measure Value. 
The in-focus lens position, then, is the one with the highest Focus Measure Value. In this manner the full cycle takes an average of 366 sec. Here, the number denotes the average time over 10 experiments with different positions of the object relative to the camera. An example of an unwanted case is when it takes more than 10 steps to get an image in focus. \emph{AF steps} denotes the average number of steps required to establish the position of the lens. \emph{Virtual pre-training} denotes that RL agent is trained with a synthetic dataset before a hardware real-world training. \emph{Deep Image Features (DF)} implies the system is universal and can be used for any kind of image due to pre-training with a large amount of data.

\paragraph*{Exposure Time} In this case, the agent only controls the $t_{ex}$ parameter of the camera. The agent interacts with the environment in same way the $A^H$ does through the Algorithm 1. A variety of light conditions is used for training, ranging from a completely dark room ($E_v$ = 13 lx) to bright lighting ($E_v$ = 300 lx). $E_v$ is adjusted every 30 min with an increment of $\delta E_v$ = 10 lx. Upon reaching the highest value, the process switches to decreasing the illuminance.

\begin{figure}
    \includegraphics[width=0.45\textwidth]{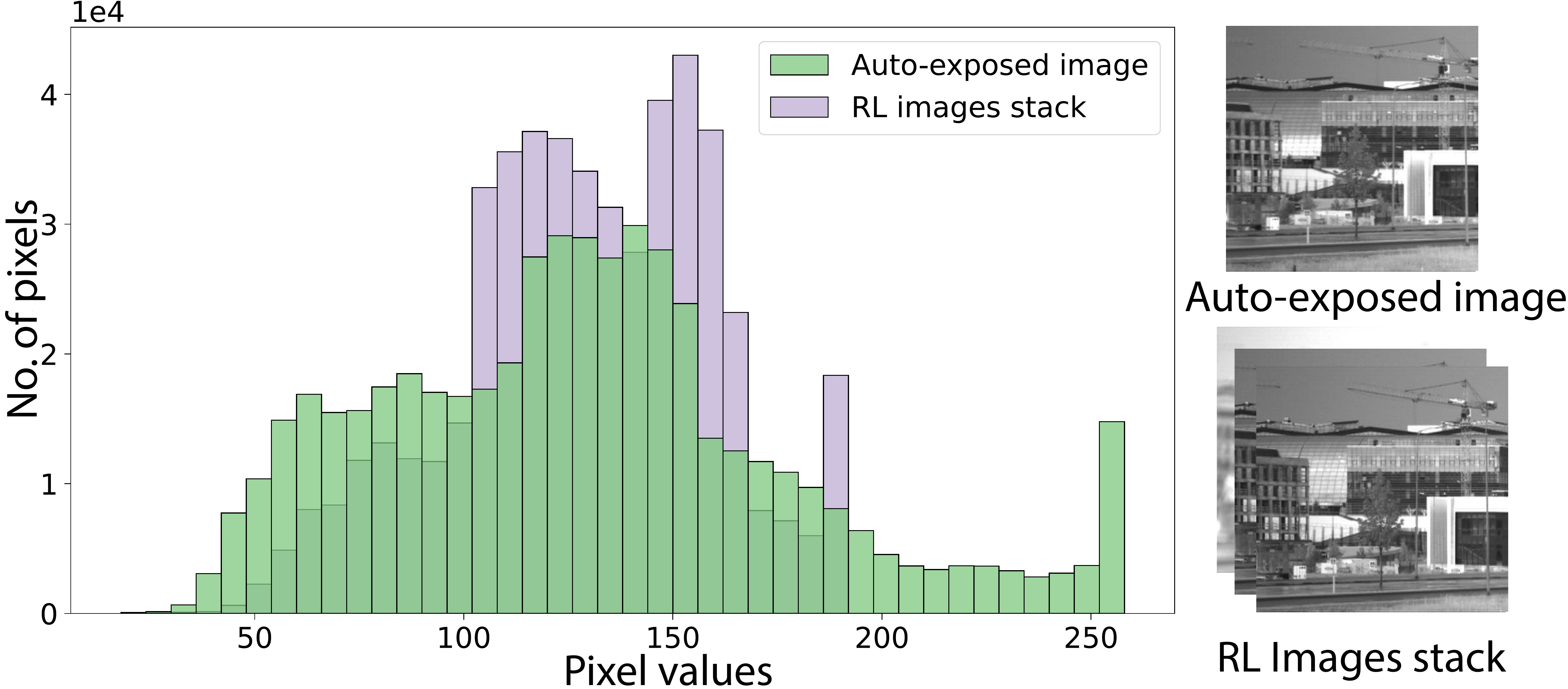}
    \caption{Histogram comparison for the RL-images stack (green) and the image obtained by the off-the-shelf algorithm embedded into the camera (purple). Taken to demonstrate consistency.}
    \label{fig:histogram}
\end{figure}

We also compared with the `brute force` AutoExposure time from the Basler camera. We confirmed that RL agent gives similar results to the native software without any exposure metering or phase difference elements (Fig.~\ref{fig:histogram}).

\begin{figure*}
    \centering
    \includegraphics[width=\textwidth]{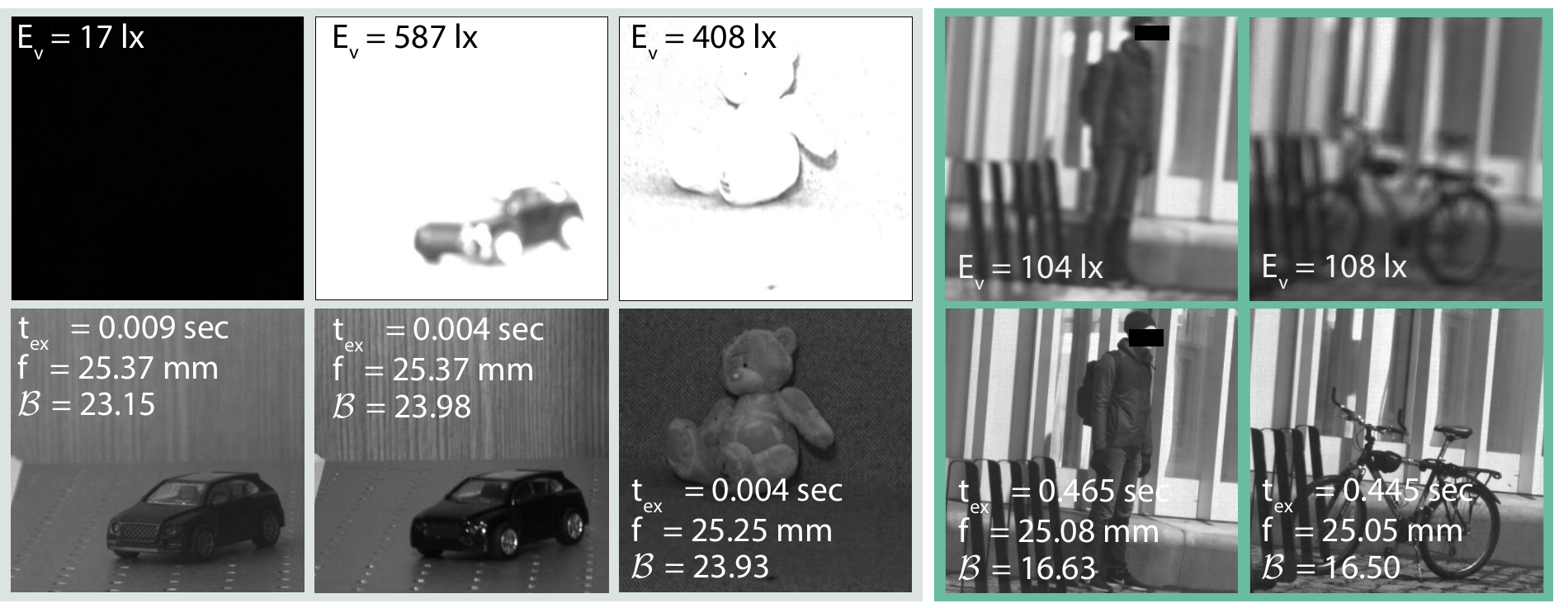}
    \caption{Automatic enhancement of poorly-lit, overexposed, and/or out-of-focus scenes by DASHA on-the-fly (inference),
    featuring objects located in the lab (left panel) and on the street (right panel). 
    \emph{First row}: examples of initial frames. \emph{Second row}: RL-optimized results. DASHA finds optimal exposure time $t_{ex}$ and focus \texttt{f} for each scene, optimizing nothing but the no-reference metric $\mathcal{B}$ computed on each frame's latent feature vector.}
    \label{fig:panel}
\end{figure*}
\begin{figure}
    \centering
    \includegraphics[width=0.9\columnwidth]{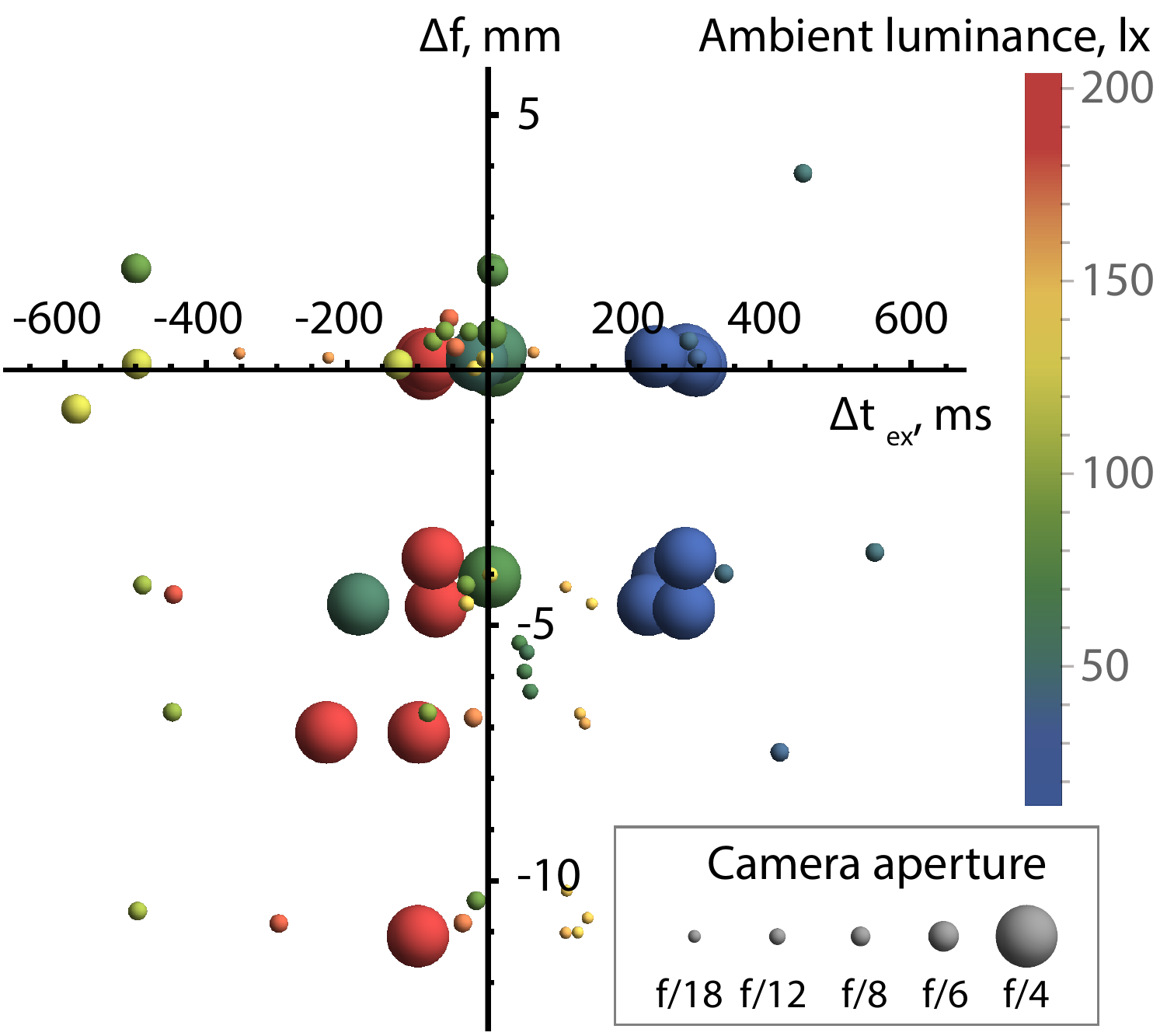}
    \caption{DASHA’s simultaneous optimization of the focal displacement of the lens (in mm) and of the exposure time (in ms). Note that the system learned to compensate for the extra light entering the frame either by subtracting some ms of exposure time or by proper re-focusing. However, given that it relies on the latent representation, in some cases it decides, \textit{e.g.}, to blur the image on purpose to make up for the missing light. These are the actions learned by DASHA which maximize the \textit{no-reference} image quality.}
    \label{fig:focal}
\end{figure}

\subsection{Multiple Agents Training}
An additional agent is introduced in order to control the $t_{ex}$. The training took place with a change in the object position (between 140 cm and 200 cm to the camera) and the background illumination (from $E_v$ = 13 lx to $E_v$ = 300 lx and backwards) every 30 minutes. We use Yolo to detect objects because of its speed. Fig.~\ref{fig:PCA} shows a summary of a large-scale study of DASHA's performance. The main components are located along the $x$, $y$, and $z$ axes (with an offset to show the difference in object detection by different algorithms). The plots show $4\times 10^3$ points, where each point corresponds to an image with a unique focal value. At different $t_{ex}$, the SOTA methods detect objects in different ways. This explains the importance of introducing an agent that controls $t_{ex}$ in the camera.

Visual results of parameters learnt by the camera and the lens are presented in Fig.~\ref{fig:panel}. The first row contains the initial images acquired by the camera at the beginning of the imaging process. The second row shows the same scenes marked as suitable according to the RL algorithm. These examples demonstrate how over-, under-exposed and blurry images are efficiently adjusted by DASHA, resulting in a successful object detection through the no-reference image quality assessment and autofocusing.
Additionally, we performed a series of inference experiments for various ambient illuminations and various apertures of the camera (Fig.~\ref{fig:focal}), confirming that the system is robust \textit{w.r.t.} different depth of the field and ambient illumination conditions.

\section{Conclusion}

In this work, we proposed a new approach to the problem of auto-tuning imaging equipment using hierarchical reinforcement learning. This methodology allows for correcting the improper focusing under various illumination conditions by employing two decentralized interacting agents that control the settings of the camera and the lens. 
The proposed way of autofocusing relies on the latent feature vector of the live image scene, being the first such method to auto-tune a camera without reference or calibration data.

The system proved efficient for detecting objects in the dark and the blurry initial states when three SOTA object detection methods originally failed to function. 
Being fast to learn (Fig.~\ref{fig:rewards_compared}), our algorithm preserves the hardware from overheating, requires no active or phase-detection elements, and is functional for a range of camera apertures (\textit{i.e.}, robust to the depth of the field variation).

Although our initial study partially explains the decisions made by DASHA (Fig.~\ref{fig:focal}), an in-depth interpretability study is to be conducted, using such feature-based feedback solutions as Grad-CAM \cite{GradCam}. Another ongoing effort will validate the performance in extreme ambient conditions (rain, dust, glare, complete darkness) and in the imaging systems with built-in active autofocusing elements (\textit{e.g.}, smartphone cameras). We envision simple integration of our framework with a wide range of consumer cameras and motorized lenses, enabling a widespread improvement both to the imaging hardware and to the embedded object detection models. 

We release our code on Github
\footnote{https://github.com/cviaai/DASHA/}.

\bibliographystyle{IEEEtran}
\bibliography{biblio}

\end{document}